\documentclass[10pt,twocolumn,letterpaper]{article}

\usepackage[pagenumbers]{cvpr} 

\definecolor{cvprblue}{rgb}{0.21,0.49,0.74}
\usepackage[pagebackref,breaklinks,colorlinks,allcolors=cvprblue]{hyperref}

\usepackage{pifont}    
\usepackage[utf8]{inputenc} 
\usepackage[T1]{fontenc}    
\usepackage{hyperref}       
\usepackage{url}            
\usepackage{booktabs}       
\usepackage{amsfonts}       
\usepackage{threeparttable} 
\usepackage{graphicx}       
\usepackage{nicefrac}       
\usepackage{microtype}      
\usepackage{xcolor}         
\usepackage{amsmath}
\usepackage{colortbl}
\usepackage{multirow}
\usepackage{enumitem}
\usepackage{wrapfig} 
\usepackage{comment}
\usepackage{algorithmic}
\usepackage{siunitx}  
\usepackage{caption}  
\usepackage{makecell} 
\usepackage[linesnumbered,ruled]{algorithm2e}
\usepackage{floatrow}  
\SetKwComment{Comment}{$\triangleright$~}{}
\makeatletter
\newcommand{\hupar}[1]{\noindent\textbf{#1}\quad}
\renewcommand{\SetKwInOut}[2]{%
  \algocf@newcommand{#1}[1]{%
    \textbf{#2:} ##1
    \par
  }
}

\def\paperID{679} 
\def\confName{CVPR}
\def\confYear{2026}

\title{Learn to See the Unseen in Low-light Spike Streams}

\author{
  Liwen Hu$^{1}$, Yang Li$^{1}$, Mianzhi Liu$^{2}$, Yijia Guo$^{1}$, Shenghao Xie$^{2}$, \\
  Ziluo Ding$^{1}$, Tiejun Huang$^{1}$, Lei Ma$^{2,*}$ \\
  \footnotesize
  $^{1}$School of Computer Science, Peking University \hspace{0.3cm} $^{2}$National Biomedical Imaging Center, Peking University \hspace{0.3cm} \texttt{lei.ma@pku.edu.cn}
}

\begin{document}
\maketitle
\begin{abstract}
Spike camera, a type of neuromorphic sensor with high-temporal resolution, shows great promise for high-speed visual tasks. Unlike traditional cameras, spike camera continuously accumulates photons and fires asynchronous spike streams. Due to unique data modality, spike streams require reconstruction methods to become perceptible to the human eye. 
However, lots of methods struggle to handle spike streams in low-light high-speed scenarios due to severe noise and sparse information. In this work, we propose \textbf{Diff-SPK}, a diffusion-based reconstruction method. Diff-SPK effectively leverages generative priors to supplement texture information under diverse low-light conditions. Specifically, it first employs an \textbf{E}nhanced \textbf{T}exture \textbf{f}rom Inter-spike \textbf{I}nterval (ETFI) to aggregate sparse information from low-light spike streams. Then, the encoded ETFI by a suitable encoder serve as the input of ControlNet for high-speed scenes generation. To improve the quality of results, we introduce an ETFI-based feature fusion module during the generation process.
Moreover, we establish the first bona fide benchmark for the reconstruction of low-light high-speed spike stream. It surpasses existing reconstruction datasets in scale and provides quantitative illumination information.
The results on multiple datasets demonstrates the superiority of Diff-SPK.
\end{abstract}    
\section{Introduction}
\label{sec:intro}

\begin{figure}[ht]
  \centering
  \includegraphics[width=\linewidth]{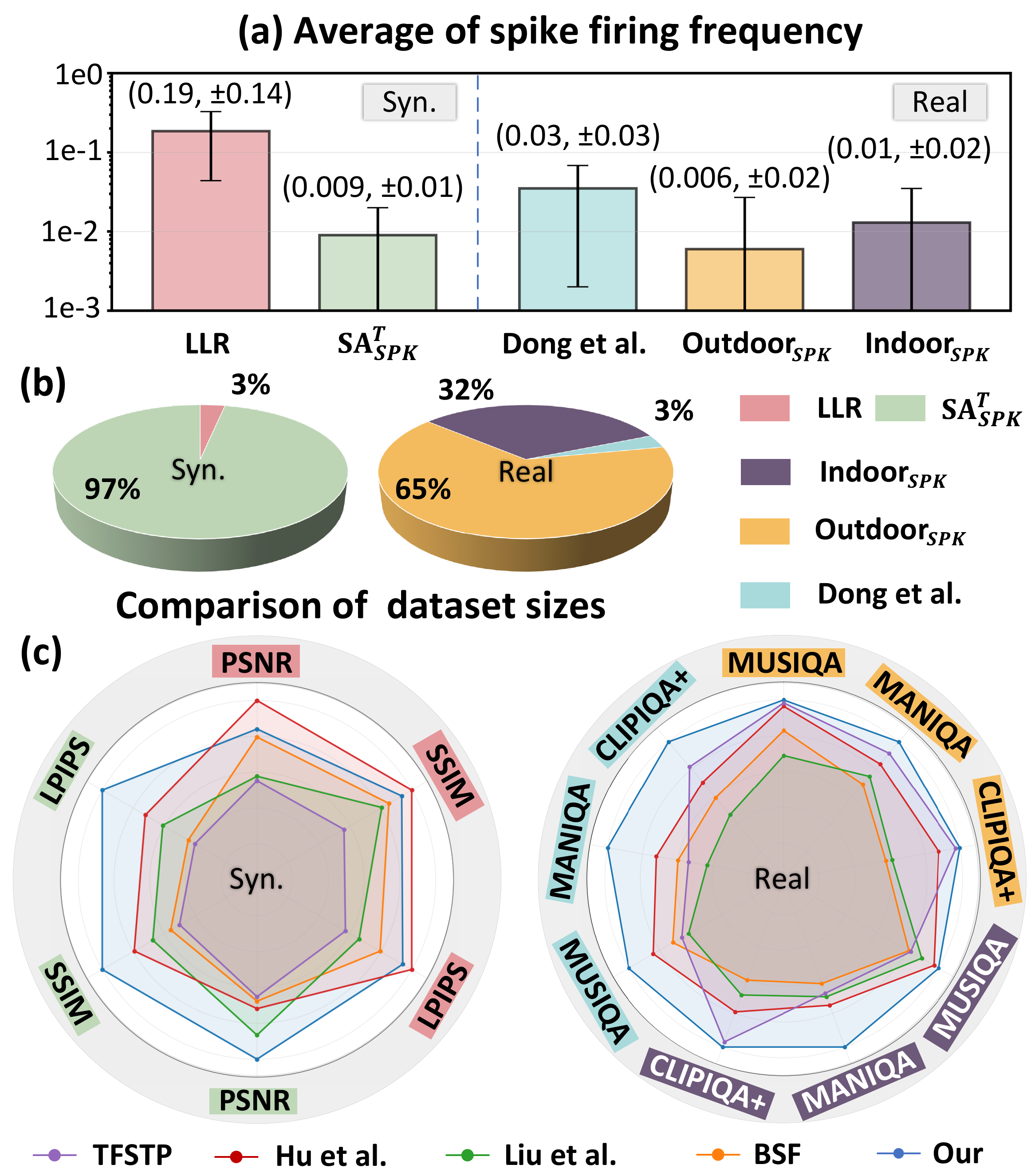}
  \caption{
  Analysis of low-light high-speed spike stream datasets and performance comparison of methods. SA$_{\textit{SPK}}^{{T}}$ (Outdoor$_{\textit{SPK}}$ $\&$ Indoor$_{\textit{SPK}}$) is proposed synthetic (real) dataset.  Details are in Sec.\ref{sec:data}. (a) Average Spike Firing Frequency (ASFF): The darker the scene, the lower the ASFF. Our datasets have darker illumination conditions compared to LLR \cite{rec8} and Dong et al. \cite{lowlighrec0}. (b) Dataset size comparison: Ours are $\sim$ \textbf{30 $\times$} larger than others. (c) Performance comparison: Our method achieves the best overall performance, demonstrating particular advantages in scenes with lower illumination. TFSTP \cite{rec2}, Hu et al. \cite{rec8}$\&$Liu et al. \cite{rec9}, and BSF \cite{rec7} represent the state-of-the-art for traditional, deep learning-based low-light, and deep learning-based normal-light reconstruction.} \label{teasor}
\end{figure}

With the development of vehicles, drones, and robotics, visual tasks in high-speed scenes have gained significant attention. Since traditional cameras often suffer from motion blur in high-speed scenes, many researchers start turning to neuromorphic sensor \cite{dvs1, dvs3,dvs5,huang20221000,spikecode}. In this context, spike camera, a type of neuromorphic sensor, shows remarkable potential in visual tasks, e.g., optical flow estimation \cite{2022scflow,zhao2022learning,flow2} and depth estimation \cite{2022spkTransformer,deep1,deep2}. Different from traditional cameras, spike camera  accumulates photons at an extremely high frame rate for each pixel, and a spike is fired once the accumulated value exceeds the fixed threshold. To perceive high-speed scenes from binary spike data, also known as spike stream, numerous reconstruction methods \cite{rec3,rec4,rec5,rec6, rec7, rec8} are proposed and achieve significant progress. 
However, as illumination intensity diminishes, vision sensors inevitably encounter a fundamental performance bottleneck \cite{dark_camera0,dark_camera1,dark_camera2, graca2023optimal}. The insufficient photon flux coupled with inherent sensor noise results in critically low signal-to-noise ratios (SNR) in captured data. Limited by early regression frameworks, e.g., CNNs \cite{cnn_framework} and RNNs \cite{rnn_framework}, these methods can only restore textures from existing data, lacking the ability to generate details. Consequently, they fail on low-light spike streams dominated by noise. Different from above frameworks, diffusion models \cite{controlnet,SD, diffusion} present a novel paradigm. By leveraging the strong generative priors, diffusion models demonstrate remarkable capability in synthesizing high-quality object textures from pure noise. A natural question arises: Can we use powerful generative models to reconstruct more texture details from sparse information of low-light high-speed spike streams?

We propose the diffusion-based reconstruction methods, Diff-SPK, to address the limitations of existing methods in low-light high-speed scenes. Diff-SPK is mainly divided into two stages. First, low-light spike stream passes through a \textbf{E}nhanced \textbf{T}exture \textbf{f}rom Inter-spike \textbf{I}nterval module (ETFI). ETFI can estimate the illumination of scenes based on spike intervals and enhance excessively dark regions into a suitable range. Next, Diff-SPK utilizes Stable Diffusion (SD) as a generative prior and finetunes a ControlNet \cite{controlnet}, where results from ETFI serve as a condition. Results from ETFI need to be encoded into the latent space before being fed to ControlNet. Diffusion models \cite{DiffBIR, supir, Pixel-Aware-diff} in image vision tasks often use VAE encoder for this operation. However, VAE encoder is not suitable for data modality of ETFI. To reduce distortion, we introduce an encoding module to extract and compress the features of input condition. Furthermore, in order to improve the guidance of input condition, i.e., ETFI, on the denoising process, a fusion module has been proposed. It can inject the information of condition into the noise latent variable during each denoising timestep.
To verify the effectiveness of Diff-SPK, we propose low-light and high-speed spike streams datasets (synthetic \& real). 
As shown in Fig.~\ref{teasor}, they have a larger scale and more extreme lighting conditions compared to the existing datasets. Our method performs well on multiple datasets and can be generalized to real low-light high-speed scenes. The contributions in this paper can be summarized as follows:

\begin{description}[leftmargin=18pt, itemindent=-5pt] 
\item[$\bullet$] We propose the diffusion-based reconstruction method for spike camera, Diff-SPK. It uses an enhanced inter-spike interval representation of spike streams, ETFI, as the  conditioning of ControlNet. ETFI can effectively aggregate sparse information in low-light spike streams.  

\item [$\bullet$] We propose a conditional encoding module and a fusion module, which can respectively compress ETFI features and enhance the guidance of ETFI on the denoising process. These modules have improved the fidelity of the reconstruction results.

\item [$\bullet$] We propose first bona fide benchmark for low-light spike streams reconstruction task. It significantly surpasses existing real low-light high-speed dataset in scale ($\sim$ \textbf{30$\times$}) and provides quantitative illumination information. Experiments demonstrate the superiority of Diff-SPK.
\end{description}

\section{Related Work}
\label{gen_inst}

\subsection{Reconstruction for Spike Camera}
\hupar{Traditional method} Early spike-based reconstruction methods, TFI and TFP \cite{spikecamera}, build on the unique statistical characteristics of spike streams. After that, TFSTP \cite{rec2} introduces neurobiological-inspired short-term plasticity to imporve reconstruction of motion region. Subsequent research by Dong et al. \cite{lowlighrec0} expand the application scope to challenging low-light, high-speed environments.

\hupar{Deep learning-based method}
To improve the performance of traditional methods, Zhao et al. \cite{rec3} introduced Spk2ImgNet (S2I), the first deep learning-based reconstruction framework.
SSML \cite{rec5} pioneers a self-supervised architecture integrating blind-spot networks, demonstrating similar performance to S2I. Concurrently, Zhang et al. \cite{rec6} develops the WGSE framework that employs hierarchical wavelet transforms for systematic noise suppression in reconstructed results. Zhao et al. \cite{rec7} further improves reconstructed results by using the framework of transformer.  Hu et al. \cite{rec8} uses RNN \cite{rnn_framework} to better handle low-light spike streams. However, the performance of the former can severely deteriorate under extreme low-light conditions. Liu et al.\cite{rec9} try to explore diffusion models for reconstruction. However, this method struggles to generalize across {high-speed} and diverse low-light conditions because its input is based on the average number of spikes in a large temporal window, introducing severe motion blur.  In this work, we present the tailored diffusion-based reconstruction method, Diff-SPK, unlocking the potential of spike cameras in low-light high-speed scenarios.



\subsection{Diffusion for Low-level Vision}
The emergence of diffusion models \cite{controlnet,SD, diffusion} has provided new insights for low-level vision tasks, such as image enhancement \cite{diff-low-light-wavelet, Diff-Retinex, LightenDiffusion}, dehazing
 \cite{Diff-Plugin, Diffusion-Models-for-Image-Dehazing}, and super-resolution \cite{Pixel-Aware-diff, supir, DiffBIR}. Recent studies leveraging powerful generative models have demonstrated remarkable performance in human visual perception. Many approaches, e.g., SUPIR \cite{supir} and DiffBIR \cite{DiffBIR}, typically employ conventional networks to obtain preliminary high-quality images, which are then refined through conditional diffusion models to improve details. They find VAE encoder particularly effective for comprehensive feature extraction and compression when high-quality images are encoded as condition. 
However, these established paradigms in above methods are inapplicable to spike streams due to data modality disparities.


\section{Preliminary}
\label{headings}

\begin{figure}[htp]
  \centering
  \includegraphics[width=\linewidth]{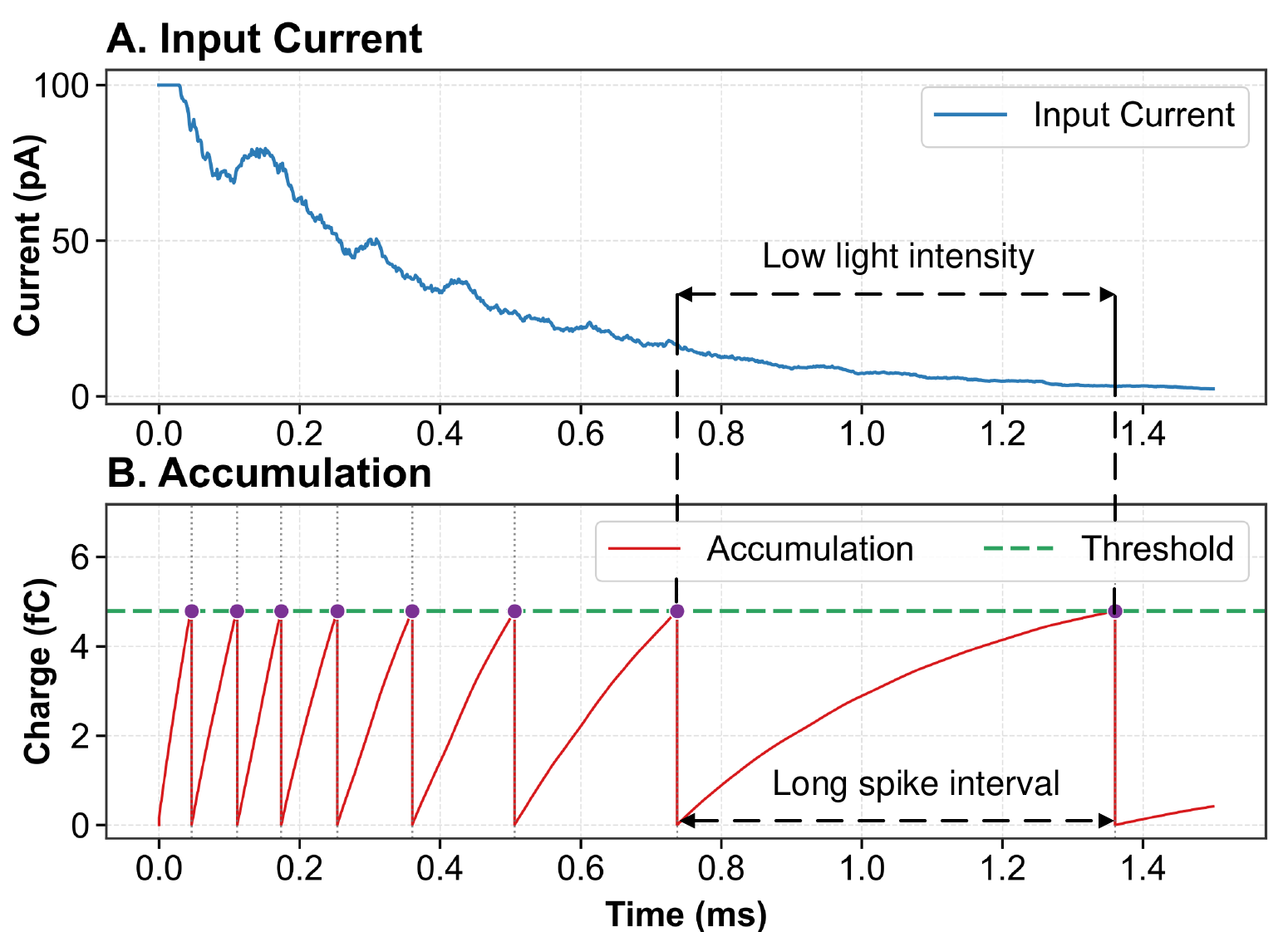}
  \caption{Illustration of spike camera model. (A) Input current. (B) Accumulation. Low-light intensity leads to long intervals between adjacent spike, i.e., sparse information.}
  \label{spike_model}
\end{figure}

\subsection{Spike Camera Model}
Spike camera \cite{huang20221000} is a kind of neuromorphic sensor with high-temporal resolution. In spike camera, each pixel initially transforms the incoming light signal into a corresponding current signal. This current signal serves as the foundational input for the sensor. As shown in Fig.~\ref{spike_model}, spike camera can continuously accumulate current.
When the accumulation at a pixel location $\boldsymbol{x} = (x, y)$ attains a predefined threshold $\phi$, a spike is fired, and the accumulation is reset. This behavior is mathematically represented as,
   \begin{align}
    &{A}(\boldsymbol{x}, t) = \int_{0}^{t} {I_{in}}(\boldsymbol{x}, \tau) d\tau\ {\rm{mod}} \; \phi, 
    \end{align}
where $A(\boldsymbol{x}, t)$ denotes the accumulated current at time $t$, and $I_{\text{in}}(\boldsymbol{x}, \tau)$ represents the input current at time $\tau$, which is directly proportional to the incident light intensity. Additionally, due to hardware constraints, the spikes are sampled at discrete intervals $nT$, where $n \in \mathbb{N}$ and $T$ is on the order of microseconds. Consequently, the output of the spike camera manifests as a binary spatio-temporal sequence $\{ S_i \mid i = 0, 1, 2, \dots \}$, with each $S_i \in \{0, 1\}^{W \times H}$ corresponding to the $i$-th spike frame. Here, $H$ and $W$ denote the resolution's height and width, respectively.
\subsection{Diffusion Model}
Diffusion models \cite{diffusion} primarily consist of a forward process and a reverse process. 
By introducing text or images during above processes, the diffusion model can produce high-quality images that adhere to the specified conditions. Our method is based on conditional diffusion models, i.e., \textbf{L}atent \textbf{D}iffusion \textbf{M}odel (LDM) \cite{SD} with \textbf{C}ontrolNet \cite{controlnet} (LDMC). LDMC is performed in the latent space. Specifically, the image $x_0$ is first encoded into the latent space as $z_0 = \mathcal{E}(x_0)$. Furthermore, in the forward process, the noisy latent variable at timestep $t$, denoted as $z_t$, can be expressed as,
\begin{equation}
    z_t = \sqrt{\bar{\alpha}_t} \, z_0 + \sqrt{1 - \bar{\alpha}_t} \, \epsilon,
\end{equation}
where $\alpha_t = 1 - \beta_t$ is the noise schedule, $\beta_t \in [0,1)$, $\bar{\alpha}_t = \prod_{s=1}^t \alpha_s$ and $\epsilon \sim \mathcal{N}(0, \mathbf{I})$ is standard Gaussian noise. 
In the reverse process, the noisy latent $z_t$ can be denoised to $z_{t-1}$ as,
\begin{equation}
    z_{t-1} = \frac{1}{\sqrt{\alpha_t}} \left( z_t - \frac{\beta_t}{\sqrt{1-\bar{\alpha}_t}} \epsilon_\theta(z_t,t, c, c_{text}) \right) + \sigma_t \epsilon,
\end{equation}
where $\epsilon_\theta(\cdot)$ is the learned noise predictor, $c$ is the condition input to ControlNet and $c_{\text{text}}$ the text prompt for LDM. 
After obtaining the clean latent $z_0$, the generated image can be decoded from  $z_0$, i.e., $y = \mathcal{D}(z_0)$. The optimization of the noise predictor is defined as follows,
\begin{equation}
\mathcal{L} = \mathbb{E}_{z_t,t,c, c_{text},\epsilon}\left[\|\epsilon_\theta(z_t, t, c, c_{text}) - \epsilon\|_2^2\right],
\label{loss}
\end{equation}
where  $t$ is randomly sampled timestep, ($z_t$, $c$, $c_{text}$) is from training data ($z_t$ is generated according to the forward process), and $\Vert \cdot \Vert_2$ denotes the 2-norm.
\section{Method}

\begin{figure*}[htp]
  \centering
  \includegraphics[width=\linewidth]{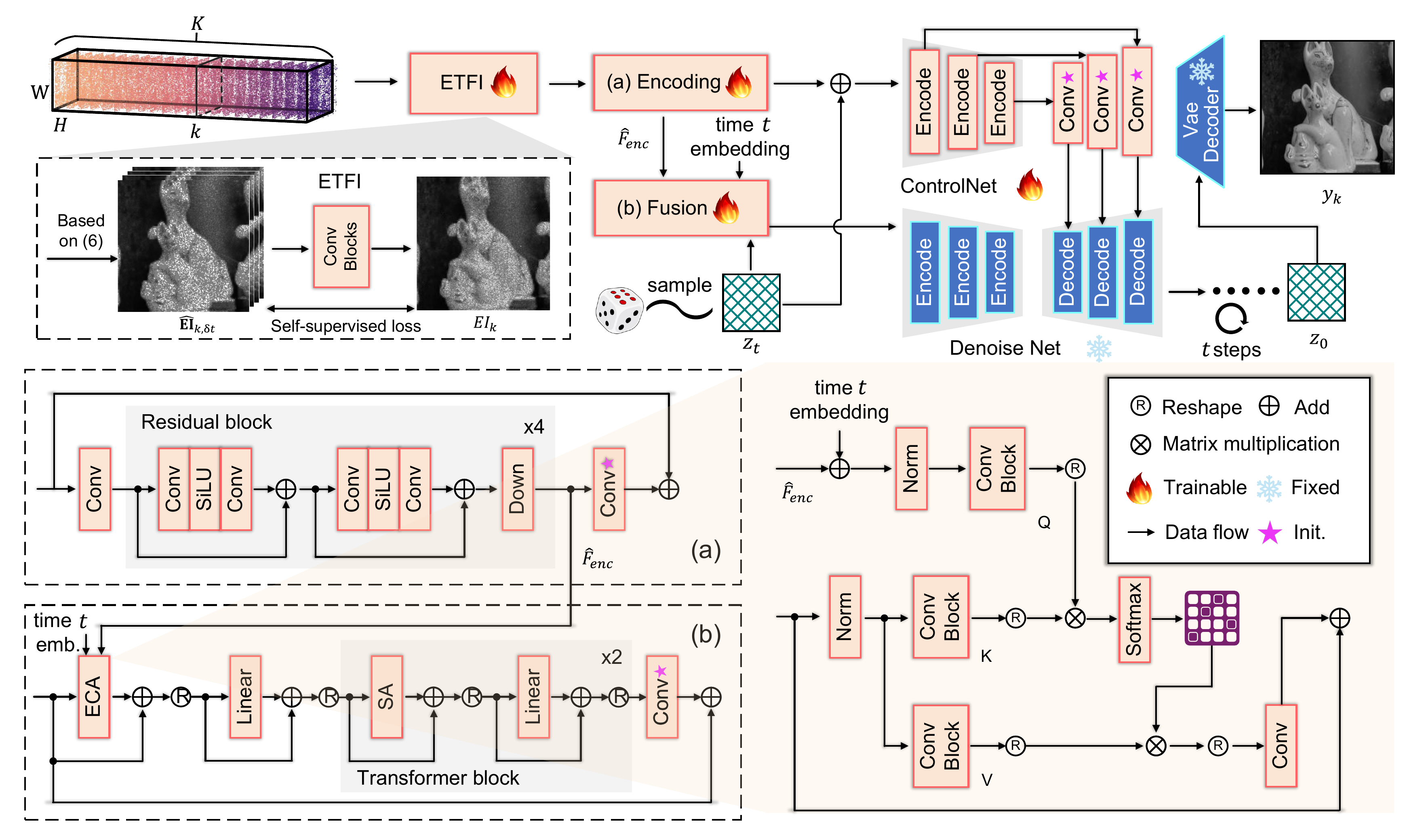}
  \caption{Framework of Diff-SPK. Diff-SPK first uses ETFI to compute and enhance light intensity information from a spike stream. The condition is through encoding (a) before being fed into ControlNet. To improve the guidance on denoising process, we introduce a fusion module (b).}
  \label{framework}
\end{figure*}

Our goal is to leverage the powerful prior of generative models for low-light high-speed spike stream  reconstruction. As shown in Fig.~\ref{framework}, we propose a tailored method, Diff-SPK, which uses the Latent Diffusion Model with ControlNet (LDMC) as the architecture. Next, we first briefly introduce the problem statement, and then elaborate on the motivation and details of the main modules including \textbf{E}nhanced \textbf{T}exture \textbf{f}rom Inter-spike \textbf{I}nterval (ETFI), condition encoding and fusion module.

\subsection{Problem Statement}
We contribute $\mathbf{S}_{k, \delta t} = \{ S_{k+i}  \in \{0, 1\}^{W \times H} \mid i \in \mathbb{Z}, -\delta t \leq i \leq \delta t \}$ as a spike stream within a temporal window $2\delta t$. The reconstruction of spike streams can be expressed as,
\begin{equation}
    y_k = \mathcal{M}(\mathbf{S}_{k, \delta t}),
\end{equation}
where $y_k$ is the light intensity information at the $k$-th frame and $\mathcal{M}$ is reconstruction model.
\subsection{ETFI}\label{sec4.2}

\hupar{Motivation} Diff-SPK builds a two-stage reconstruction pipeline. Specifically, the light intensity information of scenes is first roughly estimated as condition. Then, the fine visual results are generated by conditional diffusion models.
To obtain light intensity information, a straightforward approach is to employ a reconstruction network.
However, existing networks are difficult to reconstruct low-light spike streams, i.e., the quality of reconstruction results drops sharply once the illumination falls below a certain level. This burst quality degradation seriously hinders the learning of generative models.
We propose ETFI, a simple yet effective module, that computes and enhances light intensity information in low-light spike streams by fully exploiting temporal features. While results from ETFI remain subject to lots of quantization errors and noise, as shown in Fig.~\ref{framework}\&Fig.~\ref{vae}(a), their degradation patterns remain consistent across spike streams under different illumination conditions.  This unified degradation mode ensures that generation models can learn how to transform results of ETFI into high-quality images.


\hupar{Details} The proposed ETFI explicitly encodes and enhances light intensity information from low-light spike streams. Specifically, ETFI first uses global inter-spike intervals (ISI) to compute light intensity refer to prior methods \cite{rec8, spikecamera}
. The operation can fully extract the temporal information of spike streams. However, under  low-light illumination, some ISI are so long that the encoded light intensity cannot be perceived by people when displayed using images.
To address this, ETFI enhances the light intensity using the maximum ISI. Assume the intensity level of images is set as 256 (8 bits), and the enhanced light intensity information can be written as, 
\begin{equation}
\hat{EI_k} = \frac{max(ISI_k)}{ISI_k},
\end{equation}
where $\hat{EI_k}$ is enhanced light intensity information at the $k$-th spike frame, $ISI_k$ is a matrix composed of ISI in all pixels at the $k$-th spike frame. The advantage of this operation lies in ensuring that the minimum light intensity information is not be lost in the image. Note that the above operation may cause some pixels to be overexposed (exceeding 256), and we will further adjust the overall brightness (see appendix). Besides, to gather more temporal information and reduce quantization errors, we align the set of $\hat{EI_k}$, written as ${\hat{\mathbf{EI}}_{k, \delta t} = \{ \hat{EI_{k+i}} \mid i \in \mathbb{Z}, -\delta t \leq i \leq \delta t \}}$, to intermediate time. Specifically, we use a simple convolution block to learn this process,

\begin{equation}
\mathcal{L}_{ETFI} = \mathbb{E}_{\hat{\mathbf{EI}}_{k, \delta t}}\|\hat{\mathbf{EI}}_{k, \delta t} - ConvB(\hat{\mathbf{EI}}_{k, \delta t})  \|_2^2,
\label{loss_etfi}
\end{equation}
where $ConvB(\cdot)$ denotes convolution blocks including 5-layer convolution with 4-layer activation function and we write ${EI_k} = ConvB(\hat{\mathbf{EI}}_{k, \delta t})$ for simplicity. 
\subsection{Condition Encoding}

\begin{figure}[htp]
  \centering
  \includegraphics[width=\linewidth]{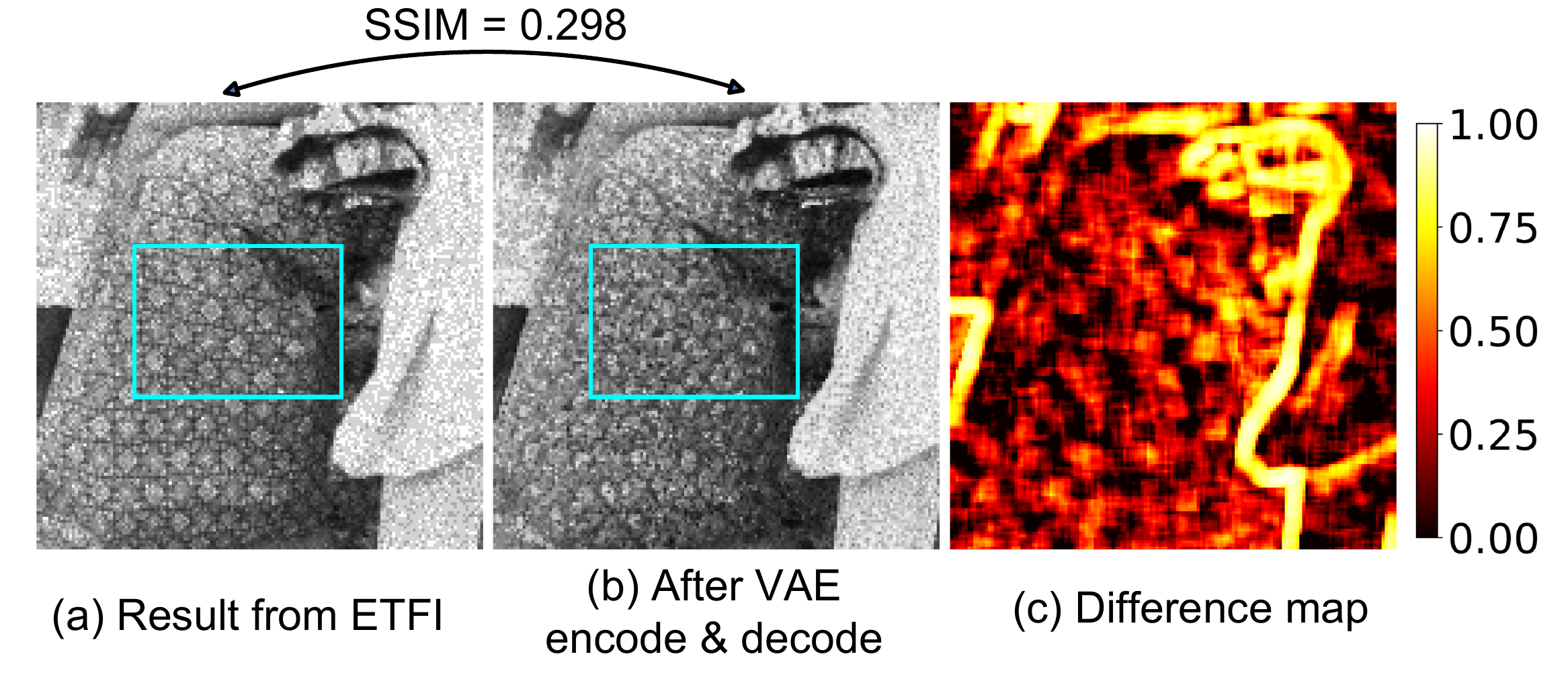}
  \caption{The influence of VAE encoder on ETFI. The difference map describes the Structural Similarity Index Measure (SSIM) of local areas between (a-b). }
  \label{vae}
\end{figure}

\textbf{Motivation} \quad Diff-SPK is based on a popular architecture, the Latent Diffusion Model with ControlNet, LDMC. For LDMC, previous work in the image vision task \cite{supir, DiffBIR,Pixel-Aware-diff} has demonstrated that the Variational Autoencoder (VAE) is well-suited to encode input images and can effectively improve generation results. However, through observation and analysis, we find that the VAE encoder are not suitable for processing results from ETFI, i.e., a large amount of spatial information is lost after encoding (see Fig.~\ref{vae}(b)). Hence, it is necessary to introduce an ETFI encoding module.


\hupar{Details} To improve fidelity of the encoded information, as shown in Fig.~\ref{framework}(a), we propose a lightweight encoding module which uses residual structures to effectively capture key information from ETFI, i.e.,
\begin{equation}
\mathcal{F}_{enc} = Conv_{init}( \underbrace{Res(Conv(EI_k))}_{\hat{\mathcal{F}}_{enc}} ) + EI_k
\end{equation}
where $Res(\cdot)$ denotes residual blocks, $Conv_{init}$ represents a convolutional layer initialized with zero weights and ${\hat{\mathcal{F}}_{enc}}$ represents the feature fed into fusion module.
Although this module is simple, it can ensure that texture details are properly retained during downsampling. 

\subsection{Fusion Module} \label{sec4.4} %


\hupar{Motivation} 
We identify that the baseline, i.e., LDMC framework suffers from fidelity degradation when ControlNet  controls the denoising process using input conditions. This is related to the control mode of ControlNet. Specifically, ControlNet first encodes the condition encoding into multi-scale features, which are then injected into the decoder of the denoising network (U-Net) via residual connections. However, as shown in Fig.~\ref{framework}, the  injection of  condition features are only at the decoder stage and it limits the guidance of condition features on the whole denoising process. To this end, we propose a fusion module to improve fidelity which can directly inject conditional features into the latent noise variable at each timestep. 


\hupar{Details} As shown in Fig.~\ref{framework}(b), we propose a fusion module to directly inject the information of condition, i.e., ETFI, into the noise latent variable $z_t$ at different timesteps. The module first employs a \textbf{E}TFI-guided \textbf{c}ross-\textbf{a}ttention (ECA) and 
a linear layer to fuse ETFI, timestep  $t$ and $z_t$ where queries are from $z_t$ and $t$, and key and value are from the feature of ETFI, ${\hat{\mathcal{F}}_{enc}}$. The process can be written as,

\begin{equation}
\mathcal{F}_{ECA} = {Linear(\underbrace{ECA(t_{emb}, {\hat{\mathcal{F}}_{enc}}, z_t)}_{\hat{\mathcal{F}}_{ECA}} + z_t)}) + {\hat{\mathcal{F}}_{ECA}}
\end{equation}
where $t_{emb}$ is the encoding of time $t$. Then, the feature $\mathcal{F}_{ECA}$ are through transformer blocks \cite{transformer} to further dig the information. Finally, fused features can be fed to U-Net for denoising. The above process can be written as,
\begin{equation}
\mathcal{F}_{fuse} = Conv_{init}(Trans( {\mathcal{F}_{ECA}})) + {z_t}
\end{equation}
where $Trans(\cdot)$ denotes transformer blocks. Furthermore, the optimization of the noise predictor in (\ref{loss}) can be rewritten as,
\begin{equation}
\mathcal{L} = \mathbb{E}\left[\|\epsilon_\theta(\mathcal{F}_{fuse}, t, \mathcal{F}_{enc}, c_{text}) - \epsilon\|_2^2\right],
\end{equation}
\section{Dataset}\label{sec:data}

Datasets are important for both evaluation (testing set) and optimization (training set) of models. As shown in Table~\ref{tab:dataset_comp}, we propose the first bona fide benchmark. Next, we will introduce these datasets respectively.

\begin{table*}[htbp]
  \floatsetup[table]{capposition=right}  
  \centering
   \resizebox{0.9\columnwidth}{!}{
  \begin{tabular}{lcrrcccc}
    \toprule
   \textbf{Name} & \textbf{Data Kind} & \multicolumn{1}{c}{\textbf{Class}} & \multicolumn{1}{c}{\textbf{Frames} / \textbf{Class}} & \textbf{Resolution} & \textbf{Motion} &\textbf{Illum.} & \textbf{GT} \\
    \midrule
    Training Dataset - SA$_{\textit{SPK}}$ & Synthetic & 111, 860 & $\approx$ 256 & (512, 512) & -  & \ding{55}& \ding{52}\\
    Testing Dataset - SA$_{\textit{SPK}}^T$ & Synthetic & 300 & $\approx$ 2,000 & (512, 512) & Random movement  & \ding{55}& \ding{52}\\
    Testing Dataset - Outdoor$_{\textit{SPK}}$ & Real & 90 & $\approx$ 20,000 & (1000, 1000) & Driving & \ding{55}& \ding{55}\\
    Testing Dataset - Indoor$_{\textit{SPK}}$ & Real & 50 & $\approx$ 20,000 & (1000, 1000) & Lens shake & \ding{52}& \ding{55}\\
    \bottomrule
  \end{tabular}%
  }
  \caption{Details of high-speed low-light spike stream datasets. Illum. (GT) denotes illumination information (scenes images).}  
  \label{tab:dataset_comp}%
\end{table*}%







\subsection{Training Dataset}
We randomly sample 111,860 images (cropped to 512$\times$512) from the SAM dataset \cite{SA} as ground truth in the training set. Further, these high-quality images are used to synthesize spike streams. The implementation comprises the following steps: (a) Degrade image quality through interpolation. (b) Simulate spike streams from the processed images. Notably, our simulation consider both the primary noise patterns of spike camera \cite{hu2024scsim, zhao2022spikingsim} and hot pixel noise. (c) Convert each spike stream into  ETFI to serve as input for Diff-SPK. See the appendix for more simulation details.

\subsection{Testing Dataset}

The captured scenarios by spike camera typically involve high-speed motion, making it infeasible to obtain clear images as ground truth (GT) for real-world datasets. Therefore, we have introduced both synthetic and real-world data as testing datasets. The synthetic data is used to evaluate the fidelity of reconstruction methods, while the real-world data is employed to evaluate the generalization of methods.

\hupar{Synthetic data} 
Beyond the training set scenarios, we randomly sample 300 scenes from the SAM dataset \cite{SA}. To simulate low-light conditions, each image was darkened by multiplying a random decay factor. For high-speed motion, we first randomly generated a motion trajectory, and then the cropping box moves along this path. We denote the dataset as SA$_{SPK}^T$. Compared to existing high-speed low-light dataset, LLR \cite{rec8}, our dataset features a larger scale and more extreme illumination conditions as shown in Fig.~\ref{teasor}.

\begin{figure}[htp]
  \centering
  \includegraphics[width=\linewidth]{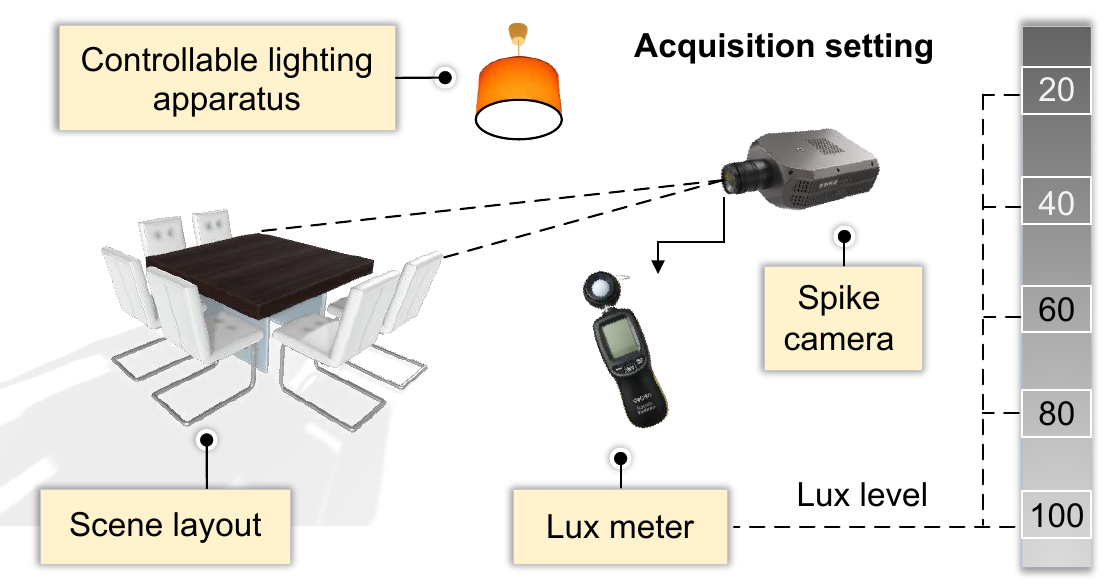}
  \caption{Instructions for obtaining illumination in Indoor$_{\textit{SPK}}$.}
  \label{data}
\end{figure}

\hupar{Real data}  We use the spike camera with 1000$\times$1000 resolution to establish a low-light high-speed benchmark, covering both outdoor and indoor scenes. For indoor scenes (Indoor$_{\textit{SPK}}$), by adjusting the light source power in Fig.~\ref{data}(b), we control spike camera to record 5$\times$10 spike streams under different illumination intensities, where 5 (10) denotes illumination levels (scene categories). Each spike stream is record for 1 second, corresponding to 20,000 frames. For outdoor scenes (Outdoor$_{\textit{SPK}}$), due to the difficulty in quantitatively controlling illumination conditions, we only collect qualitative low-light data. Specifically, spike camera is placed on a high-speed driving car to capture street scenes at dusk. The outdoor dataset contains 90 spike streams, each with a 1-second recording duration. As shown in Fig.~\ref{teasor}, the size of our datasets is $\sim$ 30 $\times$ that of the existing low-light high-speed real dataset \cite{lowlighrec0}. Moreover, our illumination conditions are much lower than \cite{lowlighrec0}.

\section{Experiment}

\subsection{Implementation detail}
Diff-SPK is based on the popular architecture, the Latent Diffusion Model with ControlNet (LDMC). We use SD1.5 \cite{SD} as our Latent Diffusion Model and ControlNet is initialized with {control\_v11f1e\_sd15\_tile.ckpt}. 
We first pre-train the ETFI module for 6,000 iterations. Then, we freeze the ETFI module and finetune ControlNet, encoding module and fusion module for 30,000 iterations (batch size=96). The learning rate is set to $10^{-5}$ and the optimizer is AdamW with default parameters.  The training process is conducted on 4 NVIDIA A800-80Gb GPUs. For inference of Diff-SPK, we use DDPM scheduler \cite{diffusion} with 50 sampling steps. We use classifier-free guidance (CFG) and CFG scale is set as 2. The positive prompt is set to empty and negative prompts are set as "low quality", "unsharp", "noisy", "weird textures", "artifacts", etc.

\subsection{Reconstruction Results}
We compare the reconstruction methods including traditional methods (TFI \cite{spikecamera}, TFP \cite{spikecamera}, TFSTP \cite{rec2}) and  state-of-the-art deep learning-based methods (SSML \cite{rec5}, S2I \cite{rec3}, WGSE \cite{rec6}, Hu et al. \cite{rec8}, BSF \cite{rec7} and Liu et al. \cite{rec9}).  These methods are evaluated on low-light high-speed datasets including real-world and synthetic data. For real-world datasets, due to the lack of GT images, we use no-reference image quality assessment (NRIQA) metrics, i.e.,  MUSIQA \cite{MUSIQ}, MANIQA \cite{MANIQA} and CLIP-IQA$^+$ \cite{CLIPiqa}, which are popular metrics in the image restoration task. For synthetic datasets, we use the reference image quality assessment metrics (PSNR, SSIM and LIPIS) and NRIQA metrics (MUSIQA, MANIQA  and CLIP-IQA$^+$). All methods are tested using the official source code and model weights. Besides, retraining on our training set do not improve baseline methods. Related details are in the appendix.

\begin{table*}[htbp]
  \centering
  \caption{Quantitative comparison on low-light high-speed datasets. \colorbox{pink!50}{Red} /  \colorbox{orange!30}{Orange} / \colorbox{yellow!30}{Yellow} indicate best / second / third performance.}
  \resizebox{0.9\columnwidth}{!}{
    \begin{tabular}{llccccccccccc}
    \toprule
    \multirow{2}[0]{*}{\textbf{Dataset}} & \multicolumn{1}{c}{\multirow{2}[0]{*}{\textbf{Metric}}} & \multicolumn{3}{c}{\textbf{Traditional Method}} & \multicolumn{7}{c}{\textbf{Deep learning-based Method}} \\     
    \cmidrule(lr){3-5} \cmidrule(lr){6-12}
     &        & \multicolumn{1}{c}{TFI} & \multicolumn{1}{c}{TFP} & \multicolumn{1}{c}{TFSTP} & \multicolumn{1}{c}{SSML} & \multicolumn{1}{c}{S2I} & \multicolumn{1}{c}{WGSE} & \multicolumn{1}{c}{BSF} & \multicolumn{1}{c}{Hu et al.} & \multicolumn{1}{c}{Liu et al.} & \multicolumn{1}{c}{Ours} \\
    \midrule
    \multirow{6}[0]{*}{LLR \cite{rec8}} & PSNR $(\uparrow)$  & 31.409 & 32.443 & 24.882 & \cellcolor{orange!30}38.432 & 36.435 & 36.094 & 35.877 & \cellcolor{pink!50}\textbf{45.075} & 26.018 & \cellcolor{yellow!30}37.898 \\
          & SSIM $(\uparrow)$  & 0.723 & 0.793 & 0.555 & \cellcolor{yellow!30}0.899 & 0.835 & 0.843 & 0.841 & \cellcolor{pink!50}\textbf{0.987} & 0.796 & \cellcolor{orange!30}0.923 \\
          & LPIPS $(\downarrow)$ & 0.360  & 0.359 & 0.441 & \cellcolor{yellow!30}0.185 & 0.231 & 0.225 & 0.222 & \cellcolor{pink!50}\textbf{0.022} & 0.355 & \cellcolor{orange!30}0.079 \\
         & MUSIQA$(\uparrow)$ & 41.445 & 35.322 & 43.667 & 41.560 & 46.171 & 44.603 &  \cellcolor{yellow!30} 49.614 &  \cellcolor{orange!30} 49.927 & 32.472 & \cellcolor{pink!50} \textbf{52.780} \\
        & MANIQA$(\uparrow)$ & 0.328 & 0.210 & 0.343 & 0.287 & 0.362 & 0.324 &  \cellcolor{yellow!30} 0.367 &  \cellcolor{orange!30} 0.371 & 0.257 & \cellcolor{pink!50} \textbf{0.391} \\
        & CLIP-IQA$^+$ $(\uparrow)$ & 0.415 & 0.304 &  \cellcolor{orange!30} 0.475 & 0.399 & 0.443 & 0.417 & 0.460 &   \cellcolor{yellow!30} 0.470 & 0.308 &\cellcolor{pink!50}  \textbf{0.513} \\
    \midrule
        \multirow{6}[0]{*}{SA$_{SPK}^T$} 
        & PSNR$(\uparrow)$ & \cellcolor{orange!30}21.909 & 19.315 & 15.315 & 16.194 & 15.043 & 17.571 & 15.898 & 16.876 & \cellcolor{yellow!30}20.346 & \cellcolor{pink!50}\textbf{23.499} \\
          & SSIM$(\uparrow)$ & 0.403 & 0.261 & 0.342 & 0.182 & 0.393 & 0.378 & 0.381 & \cellcolor{orange!30}0.542 & \cellcolor{yellow!30}0.460 & \cellcolor{pink!50}\textbf{0.683} \\
          & LPIPS $(\downarrow)$ & 0.651 & 0.785 & 0.688 & 0.710 & 0.660 & 0.642 & 0.655 & \cellcolor{orange!30}0.439 & \cellcolor{yellow!30}0.527 & \cellcolor{pink!50}\textbf{0.223} \\
        
         & MUSIQA$(\uparrow)$  & 39.090 & 39.049 & 36.175 & \cellcolor{yellow!30} 41.314 & 38.551 & \cellcolor{orange!30} 42.005 & 31.453 & 36.551 & 28.394 & \cellcolor{pink!50} \textbf{56.565} \\
          & MANIQA$(\uparrow)$  & \cellcolor{yellow!30} 0.298 &  \cellcolor{orange!30} 0.337 & 0.194 & 0.241 & 0.229 & 0.266 & 0.266 & 0.281 & 0.250 & \cellcolor{pink!50} \textbf{0.359} \\
          & CLIP-IQA$^+$ $(\uparrow)$ & 0.301 & 0.341 & \cellcolor{yellow!30} 0.368 & 0.261 & 0.246 &  \cellcolor{orange!30} 0.376 & 0.284 & 0.328 & 0.235 & \cellcolor{pink!50} \textbf{0.492} \\
    
    \midrule
        \multirow{3}{*}{Outdoor$_{\textit{SPK}}$}
      & MUSIQA$(\uparrow)$  & 42.755 & 29.314 & \cellcolor{orange!30}48.690 & 41.396 & 36.554 & 33.192 & 41.056 & \cellcolor{yellow!30}47.777 & 34.146 & \cellcolor{pink!50}\textbf{49.494} \\
      & MANIQA$(\uparrow)$  & \cellcolor{orange!30}0.377 & 0.230 & \cellcolor{yellow!30}0.374 & 0.329 & 0.269 & 0.267 & 0.281 & 0.342 & 0.305 & \cellcolor{pink!50}\textbf{0.408} \\
      & CLIP-IQA$^+$ $(\uparrow)$ & 0.402 & 0.347 & \cellcolor{orange!30}0.469 & 0.338 & 0.319 & 0.274 & 0.278 & \cellcolor{yellow!30}0.422 & 0.295 & \cellcolor{pink!50}\textbf{0.480} \\
    \midrule
    \multirow{3}{*}{Indoor$_{\textit{SPK}}$}
      & MUSIQA $(\uparrow)$ & 33.345 & 24.736 & 39.310 & 38.212 & 37.138 & 33.366 & 38.766 & \cellcolor{orange!30}46.559 & \cellcolor{yellow!30} 42.737 & \cellcolor{pink!50}\textbf{48.046} \\
      & MANIQA $(\uparrow)$ & \cellcolor{yellow!30}0.345 & 0.233 & 0.297 & 0.307 & \cellcolor{orange!30}0.362 & 0.263 & 0.272 & 0.328 & 0.306 & \cellcolor{pink!50}\textbf{0.436} \\
      & CLIP-IQA$^+$ $(\uparrow)$ & 0.390 & 0.344 & \cellcolor{orange!30}0.484 & 0.344 & 0.324 & 0.292 & 0.301 & \cellcolor{yellow!30}0.395 & 0.345 & \cellcolor{pink!50}\textbf{0.499} \\
    \midrule
    \multirow{3}{*}{Dong et al. \cite{lowlighrec0}}
      & MUSIQA $(\uparrow)$ & 35.584 & 34.338 & 36.286 & 34.461 & 37.999 & \cellcolor{yellow!30}45.440 & 39.483 & \cellcolor{orange!30}46.456 & 33.917 & \cellcolor{pink!50}\textbf{55.268} \\
      & MANIQA $(\uparrow)$ & 0.199 & 0.199 & 0.256 & 0.234 & 0.281 & \cellcolor{yellow!30}0.317 & 0.284 & \cellcolor{orange!30}0.342 & 0.206 & \cellcolor{pink!50}\textbf{0.474} \\
      & CLIP-IQA$^+$ $(\uparrow)$ & 0.248 & \cellcolor{yellow!30}0.333 & \cellcolor{orange!30}0.368 & 0.242 & 0.259 & 0.311 & 0.266 & 0.317 & 0.210 & \cellcolor{pink!50}\textbf{0.450} \\
    \bottomrule
    \end{tabular}%
  \label{tab:quant_comp}%
  }
\end{table*}%

\begin{figure*}[htp]
  \centering
  \includegraphics[width=\linewidth]{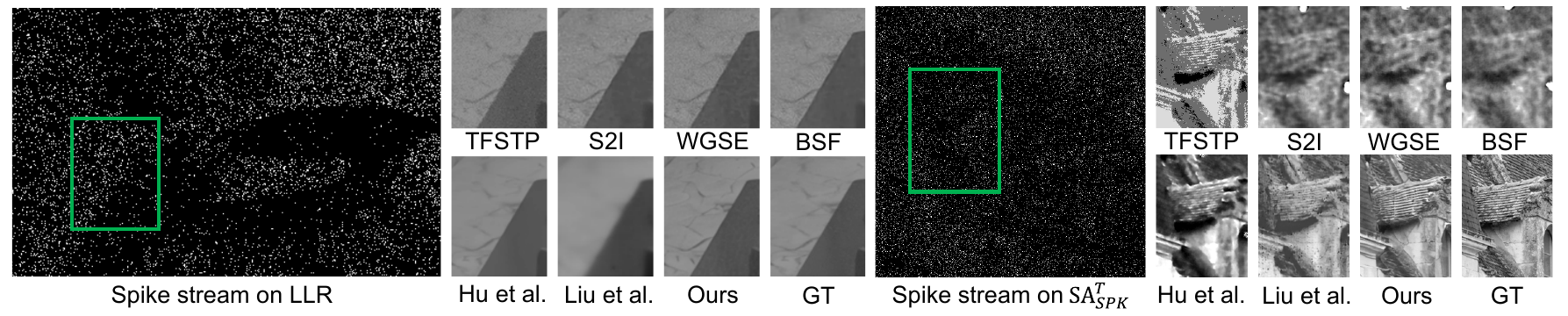}
  \caption{Visual comparison for reconstruction on LLR \cite{rec8} and SA$_{SPK}^T$. Due to low-light conditions, we perform brightness alignment between the reconstruction results and ours for viewing. \textbf{Better viewed when zoomed in.}}
  \label{experient_syn}
\end{figure*}

\begin{figure*}[htp]
  \centering
  \includegraphics[width=\linewidth]{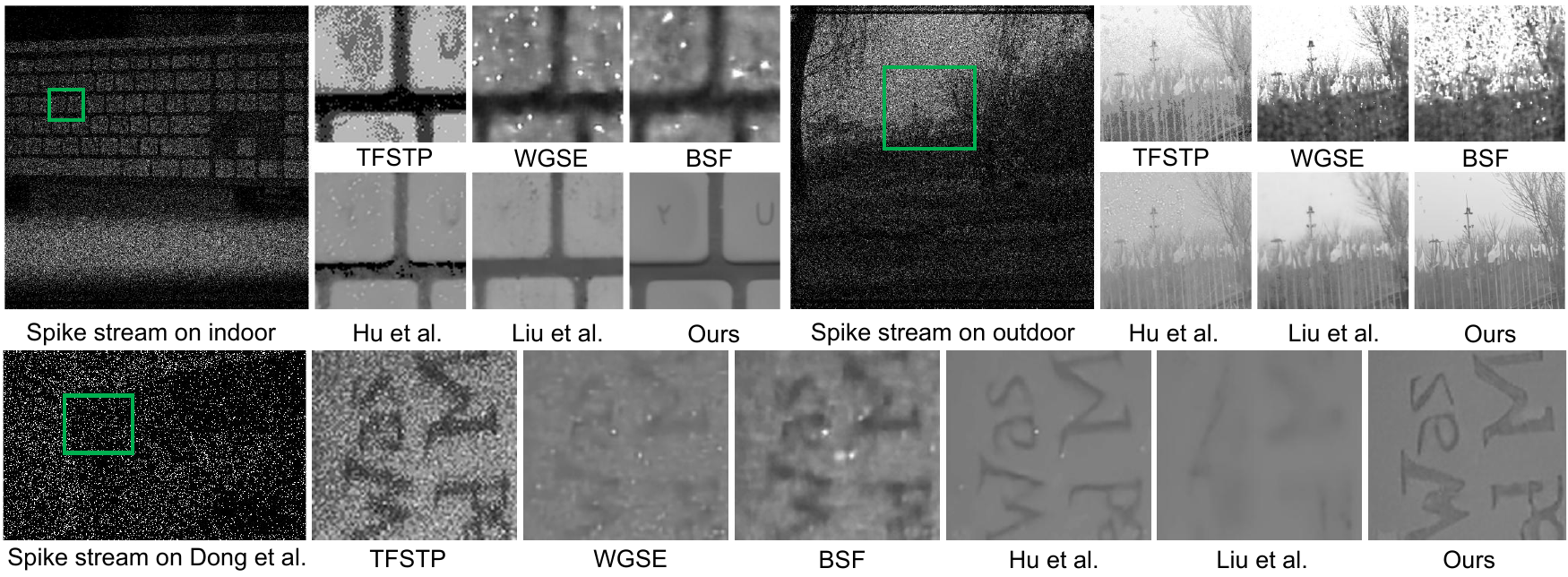}
  \caption{Visual comparison for reconstruction on Indoor$_{\textit{SPK}}$, Outdoor$_{\textit{SPK}}$, and  Dong et al. \textbf{Better viewed when zoomed in.}}
  \label{experient_real}
\end{figure*}


\hupar{Synthetic datasets} For LLR, our comprehensive performance on reference image quality assessment metrics is lower than Hu et al. \cite{rec8}. This is mainly because the overall scene of this dataset is not dark enough (see Fig.~\ref{teasor}) and some scenes are under normal-light condition. Due to the abundant information, the regression-based methods \cite{rec8, rec7, rec3, rec4, rec6} can accurately reconstruct on these normal-light data. On the contrary, the additional details generated by diffusion-based methods can reduce the fidelity to a certain extent and improve the perceived quality (Our NRIQA metrics are best).  Fig.~\ref{experient_syn} presents the visualization results. We can find Liu et al. \cite{rec9} is subjected to a large amount of motion blur, while our results are clearer. For SA$_{SPK}^T$, its scenes  are all dim and the regression-based methods are difficult to reconstruct scenes from the low-quality sparse spike streams. In this context, the fine details produced by diffusion-based methods are vital. As shown in Fig.~\ref{experient_syn}, our method has demonstrated obvious advantages.

\hupar{Real datasets} For Outdoor$_{\textit{SPK}}$\&Indoor$_{\textit{SPK}}$,  as shown in Table~\ref{tab:quant_comp}, we can find that Diff-SPK has achieved an overwhelming advantage. Fig.~\ref{experient_real} shows the reconstruction results on the two real datasets.  For Outdoor$_{\textit{SPK}}$,  our method clearly restore branches and flags. while the branches reconstructed by Liu et al. have obvious motion blur. For Indoor$_{\textit{SPK}}$,  our method detects the letters on the keyboard that are "invisible" by other state-of-the-art methods. 

For Dong et al. \cite{lowlighrec0}, the dataset was captured by Spike Camera-001T-Gen2, and it differs significantly from the spike camera used in this paper, e.g., resolution and noise level. As shown in Table~\ref{tab:quant_comp}, our method still has the best overall performance. This means that our method can be well generalized to different version of spike camera. Fig.~\ref{experient_real} shows the visualization results. We find that Liu et al.\cite{rec9}  suffers from obvious motion blur. Only our method and Hu et al. \cite{rec8} can restore all letters. Compared with Hu et al., the results of our method are clearer.

\begin{table*}[htbp]
  \centering
  \caption{Performance comparison of different methods under varying illumination conditions. The best result in each column is bolded.}
   \resizebox{1\columnwidth}{!}{
    \begin{tabular}{lccccccccc}
    \toprule
    \multicolumn{1}{c}{\multirow{2}[0]{*}{\textbf{Method}}} & \multicolumn{3}{c}{$\sim$ \textbf{40 Lux}} & \multicolumn{3}{c}{$\sim$ \textbf{60 Lux}} & \multicolumn{3}{c}{$\sim$ \textbf{80 Lux}} \\
    \cmidrule(lr){2-4} \cmidrule(lr){5-7} \cmidrule(lr){8-10}
          & \multicolumn{1}{l}{MUSIQA$(\uparrow)$} & \multicolumn{1}{l}{MANIQA$(\uparrow)$} & \multicolumn{1}{l}{CLIP-IQA$^+$$(\uparrow)$} & \multicolumn{1}{l}{MUSIQA$(\uparrow)$} & \multicolumn{1}{l}{MANIQA$(\uparrow)$} & \multicolumn{1}{l}{CLIP-IQA$^+$$(\uparrow)$} & \multicolumn{1}{l}{MUSIQA$(\uparrow)$} & \multicolumn{1}{l}{MANIQA$(\uparrow)$} & \multicolumn{1}{l}{CLIP-IQA$^+$$(\uparrow)$} \\
    \midrule
            TFI &   32.251   & 0.338       & 0.389
        & 34.702      &     0.350  &  0.396   &   36.245    &     0.358  & 0.410 \\
     TFP &   23.716   &  0.238   & 0.333
     & 23.458    &  0.234      &   0.327   &    25.915   &  0.240    &0.351  \\
    TFSTP &  38.862  &  0.287     & 
  0.490      &  36.160     &     0.271  &    0.475   &   37.494    &    0.292   &  0.479\\
    SSML  &    36.580   & 0.304      &   0.327    &    37.449   &  0.303     &    0.342   &    35.783  &   0.301    &0.343  \\
    S2I   &   36.522    &    0.257   & 0.321   &    39.465   &     0.266  & 0.317      &  30.279     &   0.266    &  0.335\\
    WGSE  & 32.502  &   0.243    &   0.286    &  35.599  &   0.277 &0.301     &     34.162   &   0.271    & 0.305 \\

    BSF   &     38.131  &  0.264     &  0.290     &  38.702  & 0.270  & 0.301      & 38.588      &  0.284     & 0.335 \\
    Hu et al. &   47.056    & 0.317     &   0.447    &  48.010 &  0.314       & 0.383      &  45.980     &   0.326   & 0.382 \\

        Liu et al. &   44.317    & 0.307     &   0.343    &  42.101 &  0.286       & 0.341      &  44.377     &   0.311   & 0.364 \\
        
    Ours   &      \textbf{48.223} &   \textbf{0.445}     &  \textbf{0.506}    &  \textbf{50.057}      &    \textbf{0.429}   &    
 \textbf{0.503}   &  \textbf{50.506}      &  
 \textbf{0.448}    &  \textbf{0.520}\\
    \bottomrule
    \end{tabular}%
    }
  \label{ab_illu}%
\end{table*}%

\begin{table*}[!h]
  \centering
  \caption{The influence of ETFI module (Baseline$_1$) and fusion module (Baseline$_2$) on the proposed datasets. Bold numbers indicate the best results in each column. All methods are trained with the \textbf{same} configuration. }
     \resizebox{1\columnwidth}{!}{
    \begin{tabular}{lccccccccc}
    \toprule
    \multicolumn{1}{c}{\multirow{2}[0]{*}{\textbf{Method}}} 
    
    & \multicolumn{3}{c}{\textbf{Indoor$_{\textit{SPK}}$}} & \multicolumn{3}{c}{\textbf{Outdoor$_{\textit{SPK}}$}} & \multicolumn{3}{c}{\textbf{SA$_{\textit{SPK}}^{{T}}$}} \\
        \cmidrule(lr){2-4} \cmidrule(lr){5-7} \cmidrule(lr){8-10}
          & \multicolumn{1}{l}{MUSIQA$(\uparrow)$} & \multicolumn{1}{l}{MANIQA$(\uparrow)$} & \multicolumn{1}{l}{CLIP-IQA$^+$$(\uparrow)$} & \multicolumn{1}{l}{MUSIQA$(\uparrow)$} & \multicolumn{1}{l}{MANIQA$(\uparrow)$} & \multicolumn{1}{l}{CLIP-IQA$^+$$(\uparrow)$} & \multicolumn{1}{l}{PSNR$(\uparrow)$} & \multicolumn{1}{l}{SSIM$(\uparrow)$} & \multicolumn{1}{l}{LPIPS$(\downarrow)$} \\
    \midrule
        Baseline$_1$ & 46.268& 0.375 & 0.430  & \textbf{50.895} & 0.316 & 0.411 & 15.605 & 0.429 & 0.474 \\
    Baseline$_2$ & 47.867 & 0.408  & 0.471 & 48.955 & 0.381 & 0.451 & 23.365 & 0.679 & 0.233 \\
    Ours   & \textbf{48.046} & \textbf{0.436} & \textbf{0.499} &  48.494 & \textbf{0.408} & \textbf{0.480} & \textbf{23.499} & \textbf{0.683} & \textbf{0.223} \\
    \bottomrule
    \end{tabular}%
    }
  \label{ab_model}%
\end{table*}%

\subsection{Ablation}


\hupar{Different illumination}  We would like to emphasize that Diff-SPK does not excel only under specific low-light conditions, which might otherwise lead to inflated average quantitative results of Indoor$_{\textit{SPK}}$ in Table~\ref{tab:quant_comp}. In fact, Diff-SPK demonstrates superior reconstruction capabilities across general low-light conditions compared to existing methods. Table~\ref{ab_illu} presents quantitative results under three distinct illumination levels (40, 60 and 80 Lux) in Indoor$_{\textit{SPK}}$. 

\hupar{ETFI module} 
As mentioned in Sec.~\ref{sec4.2} (motivation), we study the baseline, i.e., first applying a state-of-the-art reconstruction method \cite{rec8} of low-light high-speed spike streams to obtain an initial image, followed by refinement using Diff-SPK. The results, denoted as Baseline$_1$ in Table~\ref{ab_model}, demonstrate that employing the reconstructed results as the condition leads to a performance degradation. This is because the method shows highly irregular degradation patterns under different low-light conditions, which adversely affects Diff-SPK's inference capability.



\hupar{Fusion module} We investigate the impact of the fusion module on Diff-SPK. As shown in Table~\ref{ab_model} (Baseline$_2$), Diff-SPK without the fusion module inferior performance compared to our full method. As discussed in Sec.~\ref{sec4.4} (motivation), this performance gap stems from the fact that ControlNet only injects conditional information, i.e., ETFI, in the decoding stages of the U-Net, thereby weakening the guidance of ETFI throughout the denoising process.

\vspace{-1pt}
\section{Conclusion}
We present \textbf{Diff-SPK}, a diffusion-based framework specifically designed for the reconstruction of low-light spike streams. Combined with large pre-trained generative models, our method effectively perceives and enhances details under different low-light conditions. Besides, we establish the first bona fide benchmark for low-light high-speed spike stream reconstruction which has larger scale than exist real dataset ($\sim$ \textbf{30$\times$}). Experiments show that Diff-SPK can recover fine-grained texture which is invisible for other methods.

\vspace{-2pt}
{
    \small
    \bibliographystyle{ieeenat_fullname}
    \bibliography{main}
}


\end{document}